\documentclass[letterpaper, 10 pt, conference]{ieeeconf}  

\usepackage{gensymb}
\usepackage{graphicx}
\usepackage{float}
\usepackage{url}

\IEEEoverridecommandlockouts                              

\usepackage{makecell}

\title{\LARGE \bf
Real-time Detection, Tracking, and Classification of Moving and Stationary Objects using Multiple Fisheye Images
}

\author{Iljoo Baek$^{*}$, Albert Davies$^{*}$, Geng Yan, and Ragunathan (Raj) Rajkumar
\thanks{*Both authors contributed equally to this manuscript.}
\thanks{Iljoo Baek and Albert Davies are with the Department of Electrical and Computer Engineering, Carnegie Mellon University, Pittsburgh, PA 15213, USA ibaek, albertd@andrew.cmu.edu}%
\thanks{Geng Yan is with Robotics Institute, Carnegie Mellon University, Pittsburgh, PA 15213, USA gyan@andrew.cmu.edu}%
\thanks{Raj Rajkumar is with the faculty of the Department
of Electrical and Computer Engineering and Robotics Institute, Carnegie Mellon University, Pittsburgh, PA 15213, USA}%
}

\begin{document}

\maketitle
\thispagestyle{empty}
\pagestyle{empty}

\begin{abstract}

The ability to detect pedestrians and other moving objects is crucial for an autonomous vehicle. This must be done in real-time with minimum system overhead. This paper discusses the implementation of a surround view system to identify moving as well as static objects that are close to the ego vehicle. The algorithm works on 4 views captured by fisheye cameras which are merged into a single frame. The moving object detection and tracking solution uses minimal system overhead to isolate regions of interest (ROIs) containing moving objects. These ROIs are then analyzed using a deep neural network (DNN) to categorize the moving object. With deployment and testing on a real car in urban environments, we have demonstrated the practical feasibility of the solution\footnote{The video demos of our algorithm have been uploaded to Youtube: \url{https://youtu.be/vpoCfC724iA}, \url{https://youtu.be/2X4aqH2bMBs}}.

\end{abstract}

\section{INTRODUCTION}

Autonomous cars are currently primed for mass adoption by the consumer market. The recent slew of announcements by several car makers and automotive suppliers indicate that we continue on the path to having autonomous vehicles cohabiting our streets with other objects. Some examples of these objects include pedestrians, vehicles, strollers and shopping carts to name a few. As opposed to most other consumer technologies, autonomous vehicles behaving erroneously can result in significant harm to humans and property. Therefore, the ability to detect and recognize various objects around the car is paramount to ensuring safe operation of the vehicle. In this paper, we present a solution to detect and classify several moving objects around the vehicle in real time. This solution was developed for use in Carnegie Mellon University's autonomous driving research vehicle, Cadillac SRX~\cite{SRX}.

\subsection{Related work}

Several approaches to detecting obstacles around the car have been explored. The information about objects around the car can be obtained through various sensors. These include LIDAR and RADAR-based approaches used by~\cite{wang2017pedestrian}. Others include stereo camera-based approaches to make use of the additional depth information available~\cite{stereo}. Bertozzi et al.~\cite{bertozzi2015360} use images from 4 fisheye cameras to detect and track objects. Another approach is to combine the data from various sensors to arrive at a more accurate estimate. This sensor fusion-based approach has also been explored for Carnegie Mellon University's  autonomous driving research vehicle~\cite{SRX}\cite{cho2014multi}. Once the information about the environment has been obtained, the detection of objects can be done by processing one single frame at a time or by comparing the images from adjacent frames. Detection using processing from a single frame can be done using feature-based methods or machine learning. These methods analyze a single frame to find various object classes like pedestrians and cyclists. The disadvantage of this approach is that they impose high computational demand and they can also miss objects that can be of potential danger but were not known a priori while training the model. Detection using processing of adjacent frames takes advantage of object motion to reduce the search space and computation time. Optical flow-based methods are the most popular for this latter approach. When combined with tracking, these methods can detect objects even in the absence of relative motion, if the object had been moving anytime in the past. Yet others combine many of these methods to develop a hybrid approach~\cite{choi2012realtime}.

\begin{figure}[t]
  \centering
    \vspace{0.1in}
	\includegraphics[width=1\columnwidth]{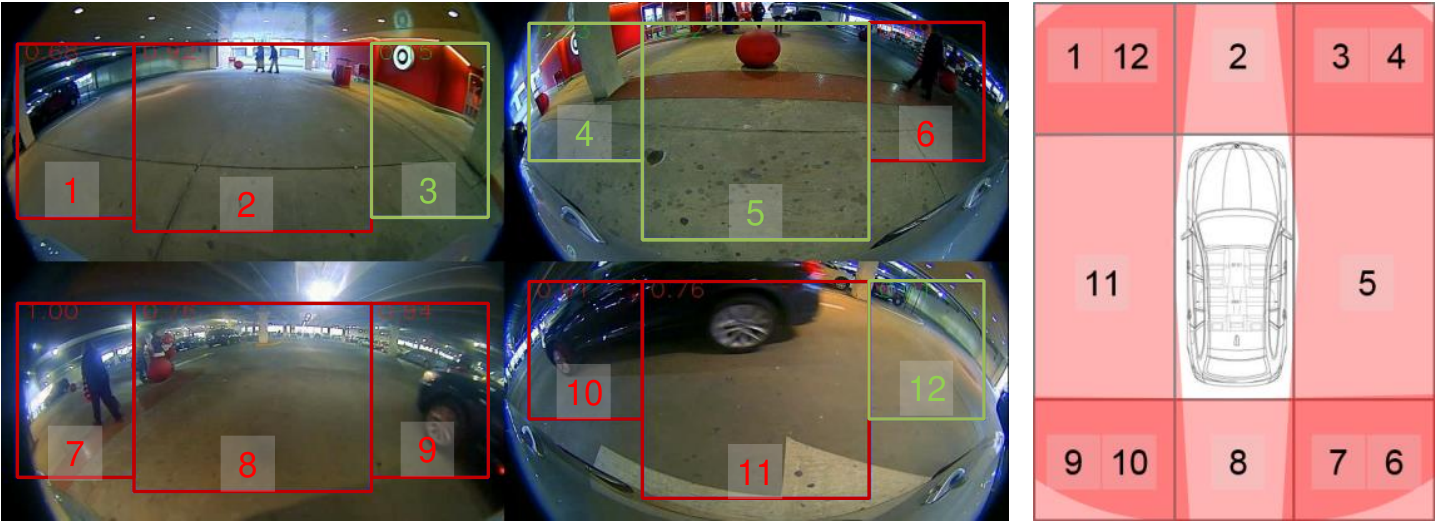}
  \caption{Left: Input image configuration, Anti-Clockwise from top left- front, back, left and right cameras respectively. The rectangle boxes indicate the 12 regions of interest. Right: Mapping of the ROIs to regions around the vehicle}
  \label{fig:ROI}
\end{figure}

\subsection{Our Contributions}

\begin{figure*}[]
	\includegraphics[width=0.96\linewidth]{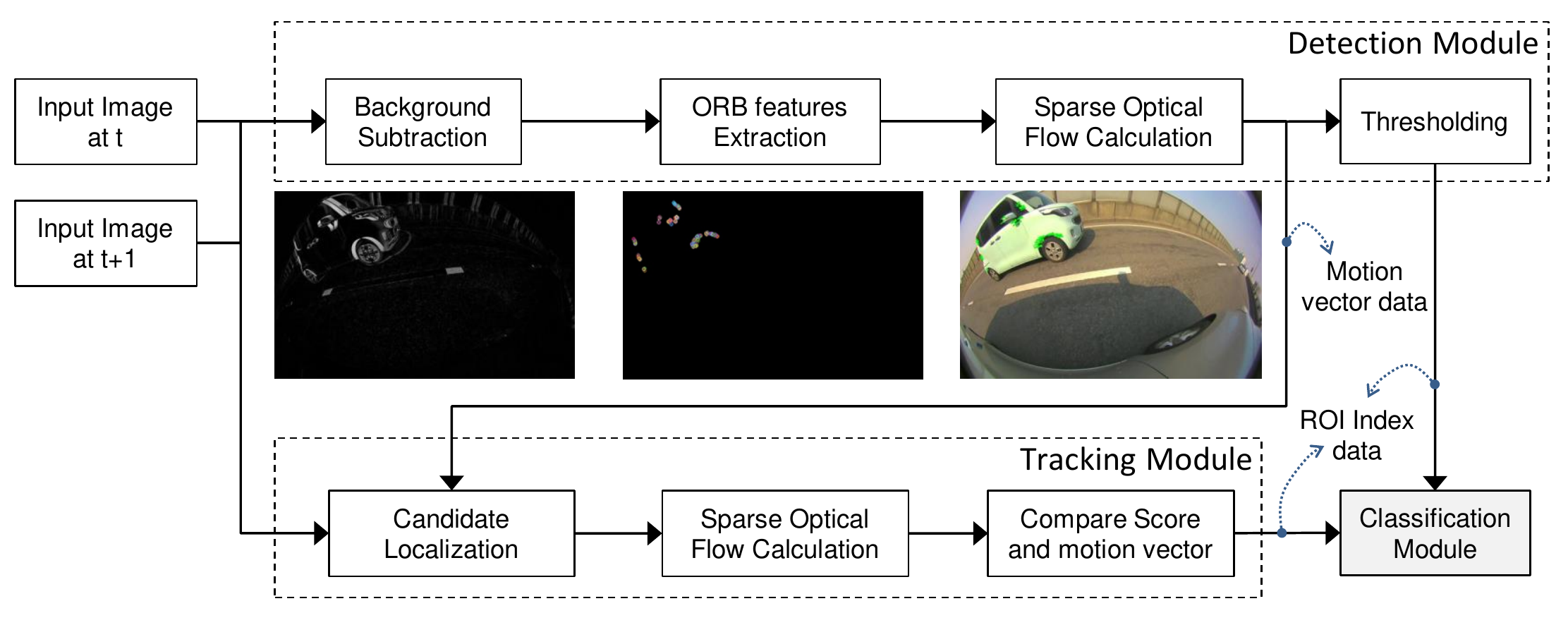}
	\caption{A schematic overview of our detection and tracking module: The detection module localizes potential moving objects in images at time t and time t+1 and calculates motion vectors. Predefined ROIs are selected from the motion vector information. The image at time t+1, selected ROIs, and the motion vector data are given as inputs to our tracking module. The feature points are detected in the t+1 frame with high confidence and the resultant motion vector for the feature points is small, we conclude that the moving object is still present in the ROI but has stopped moving. Our classifier estimates the target classes of the moving/static objects from either our detection or tracking.}
	\label{fig:detection_tracking_module}	
\end{figure*}

An autonomous vehicle must be capable of driving at different speeds based on the surrounding environment and operating conditions. In specific, its speed can range from zero, close to zero, all the way to highway speeds. We distinguish between two operating regimes based on whether the autonomous vehicle can stop rapidly and safely, or it needs to travel a considerable distance before coming to a safe and comfortable halt. These two operating modes lead to very different design considerations. We have developed our solution to work in the former operating mode of driving at low speeds and hence being able to stop quickly yet safely. In this mode, it is important to detect whether an obstacle is coarse-grained present in regions around the car. Once an obstacle is detected in any of the surrounding regions, the car can plan a path to avoid hitting the obstacle. In the worst case, the vehicle can wait until the region is considered safe to move into. Therefore, we do not need to detect the exact bounding box or distance of the obstacles relative to the vehicle.

We developed a solution to fit the practical need for having the solution perform accurately in real-time on the vehicle. We made several design choices to meet our objectives. The moving object detector is run on the CPU, while the DNN-based classifier is run on a GPU. One of the major challenges is to maintain a low computation overhead. This is critical to achieve real-time processing of simultaneous streams from 4 fisheye cameras. To achieve such a low computation overhead, we eschewed the De-Warping stage that similar solutions use~\cite{bertozzi2015360}. Our algorithm was developed to work directly on the raw images from the fisheye cameras. Instead of processing the 4 video streams individually, we merge the four video streams into a single video sequence and process them as a single stream. We also deal with a fixed region of interest in the 4 video streams as shown in Figure~\ref{fig:ROI}. Through experimentation with video streams captured from real-world conditions, we found that these 12 ROIs work best to detect presence of obstacles in the 8 regions around the car. This enabled us to narrow down the search space and transform the task of finding where the moving object is into a simpler one of figuring out whether a particular region of interest contains a moving object. For ensuring that a previously detected moving object continues to be detected even after it stops moving, we have developed a simpler version of tracking using the sparse Lucas Kanade algorithm.

Another priority of ours was that our solution must be portable enough to be platform-independent. The detector and tracking modules were implemented in OpenCV to achieve this objective. The modules were also developed to be independent from each other. This allows us to enable only a subset of the modules as necessary. This also allows the modules to be integrated without much effort with newer solutions. We have done extensive testing of the entire solution on a real car as well as using offline computation. Our entire solution has been ported to x86, TI TDA2x, and NVIDIA TX2. Performance and accuracy measurements were taken for all these system implementations.

\begin{figure*}[]
    \centering
	\includegraphics[width=0.96\linewidth]{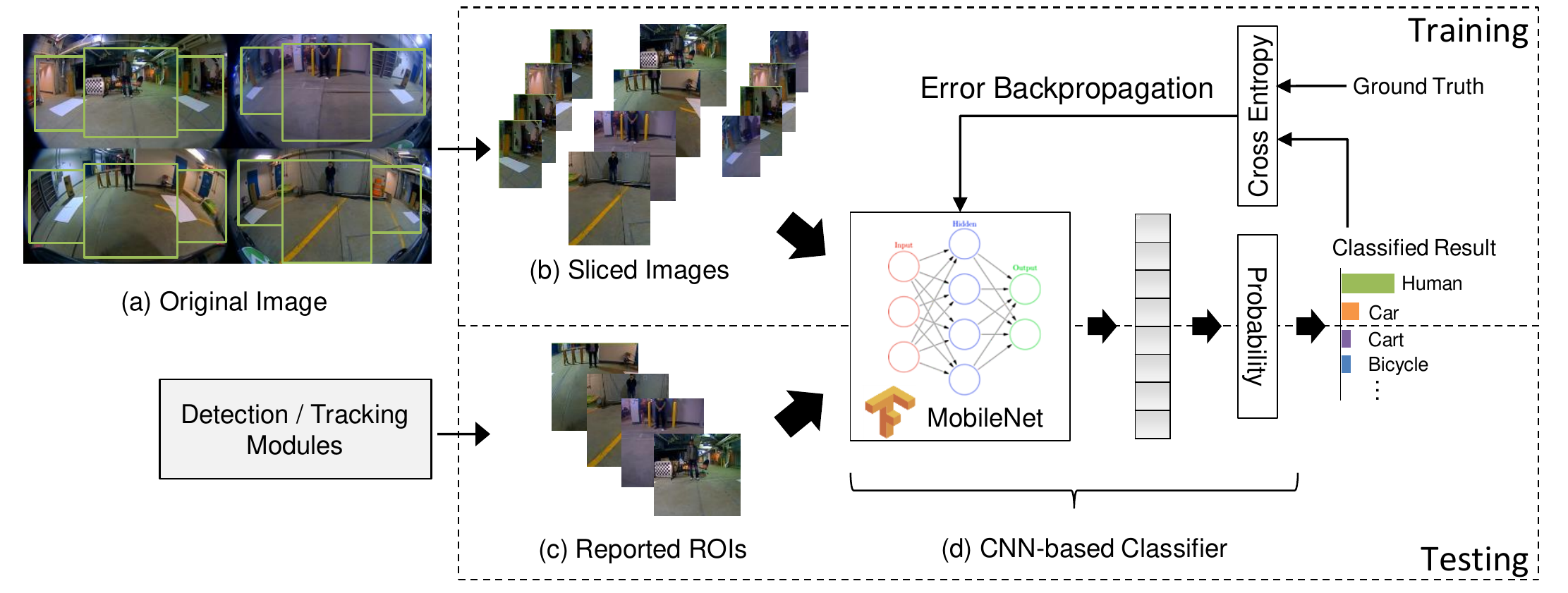}
	\caption{A schematic overview of our classification module: (a) Green rectangles represent fixed ROIs. Each ROI can be selected differently according to vehicle body and nearby environment. (b) The original image is split into individual images with class label and the sliced images are re-sized into 224x224. (c) Either all ROIs or selected ROIs from the detection and tracking module can be pass to MobileNet. (d) 4096 vector descriptor is created as an output. SoftMax with Cross Entropy provides a probability distribution over K categories. }
	\label{fig:classification_module}	
\end{figure*}

\section{Architecture and Algorithms}

The overall architecture of our solution includes detection, tracking and classification modules. The detection module detects if there is a moving object in any of the regions of interest. The tracking module is responsible for ensuring that a previously-detected object continues to be detected even if it stops moving. The classification module uses deep learning to categorize the moving object. The function and design of these modules are explained in the following sections.

\subsection{Detection of Moving Objects}

The detection module detects if there is a moving object in any of the regions of interest. The different regions of interest were chosen to align with practical needs to indicate vehicle and other objects approaching from the left, right or center of each field of view. 

Figure~\ref{fig:detection_tracking_module} illustrates the pipeline of our detection and tracking modules. As shown, the first stage in detection is the background subtraction of adjacent frames. This narrows our search space to only the pixels where we believe motion to be present. In this background-subtracted image, we find the appropriate feature locations to track. Similar solutions use various feature descriptors such as SIFT~\cite{sift}, SURF~\cite{surf}, BRIEF~\cite{brief} and ORB~\cite{orb}. 
We decided to use the ORB feature descriptor chiefly because of its efficiency of computation. Once the feature points are extracted, we use the sparse Lucas Kanade algorithm~\cite{baker2004lucas} to find the corresponding feature points in the subsequent frame. The vectors connecting the feature points from the previous frame to the corresponding ones in the current frame were summed for every region of interest separately. The length of the resultant vector was thresholded to determine if the region of interest contains any moving object.

\subsection{Tracking of Stopped Objects}

With only detection, there is the drawback that a moving object will stop being detected if the object enters a region of interest and then stops moving. This behavior is not what we intend. We would like to know of every potential object around the car that can pose danger, even though it is not moving currently. Hence, we introduced a tracking module that would track a previously-detected object even after it stops moving. Tracking of moving objects in a video stream is a well-studied problem with many state-of-the-art trackers available~\cite{kristan2015visual}. Most state-of-the-art trackers perform pretty well with high sensitivity and specificity. However, the computation overhead of these trackers is high. Since we are concerned with detecting the presence of moving objects in the regions of interest, our requirements are less demanding than that offered by these turn-key solutions. We need not track multiple objects, since we only generate warnings for the presence or absence of moving objects in each ROI. We also did not need to track the exact bounding boxes for the objects. We leverage these looser requirements to obtain lower computation overhead. In our solution, once we detect the presence of moving objects, we store those feature points. When the detector transitions from positive to negative, we run another iteration of the sparse Lucas Kanade algorithm on the stored feature points from the previous frame. If the feature points are detected in the current frame with high confidence and the resultant motion vector for the feature points is small, we conclude that the moving object is still present in the ROI but has stopped moving. This ROI is considered to contain the moving object until the detector module kicks in and freshly detects a moving object, at which point, the stored features are discarded.

\subsection{ROI Classification}

The detector and tracking modules are used to find the regions of interest with moving objects. However, to make meaningful use of this information, the system needs to know the nature of the moving object. Specifically, we would like to categorize the moving object as pedestrian, bicycle, shopping cart, etc. This information can then be used by the autonomous vehicle software to take action tailored to the category of the detected object. For this purpose, we introduce a deep neural network-based classification module.

\subsubsection{MobileNet}
Since AlexNet~\cite{alexnet} in 2012, Convolutional Neural Networks have proved to be the state-of-the-art methodology in Computer Vision. It has outperformed traditional approaches in various tasks, such as recognition, segmentation, and detection. Over time, CNNs in use today have grown to be very deep and complex. This has led to an increase in accuracy. However, these networks are also large in size and their response times are slow. In many real-world applications like our system, the model needs to execute on a small, low-power embedded system without a powerful GPU to provide sufficient processing power to run a computationally heavy deepnet in real-time. In 2017, Andrew et al.~\cite{howard2017mobilenets} proposed \textit{MobileNet}, which substitutes the standard convolution layers with depth-wise convolution layers and point-wise convolution layers to achieve small, low-latency models that meet the requirement for our application.

MobileNet factorizes a standard convolution into a depth-wise convolution and a point-wise convolution, which is referred to as depth-wise separable convolution. The depth-wise convolutions apply a single filter for each input channel. Then, the output map is fed into a $1 \times 1 \times output\_size$ convolution to combine the output of the depth-wise convolution layer. The factorization actually splits a standard one-step convolution operation into 2 separate steps. Because the separation restricts space requirements and simplifies the computation, depth-wise separable convolution can achieve a reduction in computation of $$\frac{1}{N} + \frac{1}{D^2_K}.$$

%
%

According to~\cite{howard2017mobilenets}, MobileNet uses $3 \times 3 $ depth-wise separable convolutions resulting between $8$ to $9$ times less computation than the use of standard convolution with a reduction in accuracy of only about $1 \%$.

\begin{figure}[t]
    \centering
	\includegraphics[width=0.7\columnwidth]{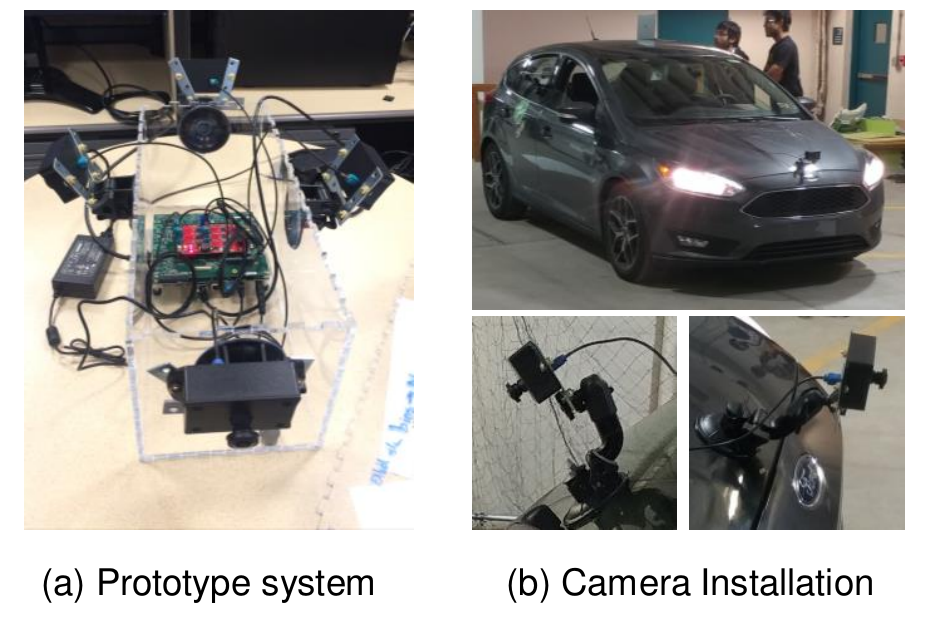}
	\caption{Surround View System for capturing data: TI-TDA2x evaluation board with four 190\degree FOV cameras.}
	\label{fig:surroundview}	
\end{figure}

\subsubsection{Classifier}
MobileNet provided us features extracted from images as descriptors. We utilized these descriptors to build a classifier. Here, we chose to use SoftMax together with Cross Entropy as the loss function.

SoftMax is normally used after the final layer to map real values to the interval $[0,1]$, in order to represent a probability distribution over $K$ possible categories. Then, we use cross entropy to compute loss and provide a gradient for the back-propagation phase. Before training, we first obtain the frames from the video, and then split them into regions of interest.  Because MobileNet takes input images of dimension $224 \times 224,$ we re-sized all different-resolution images into $224 \times 224$ images and labeled them as shown in Figure~\ref{fig:classification_module}.

\section{Evaluation}
This section describes our experiments measuring the performance of our moving-object detection system using different hardware platforms and images acquired under various areas.

\subsection{Hardware and Implementation}

\begin{figure}[t]
	\centering
		\includegraphics[width=\columnwidth]{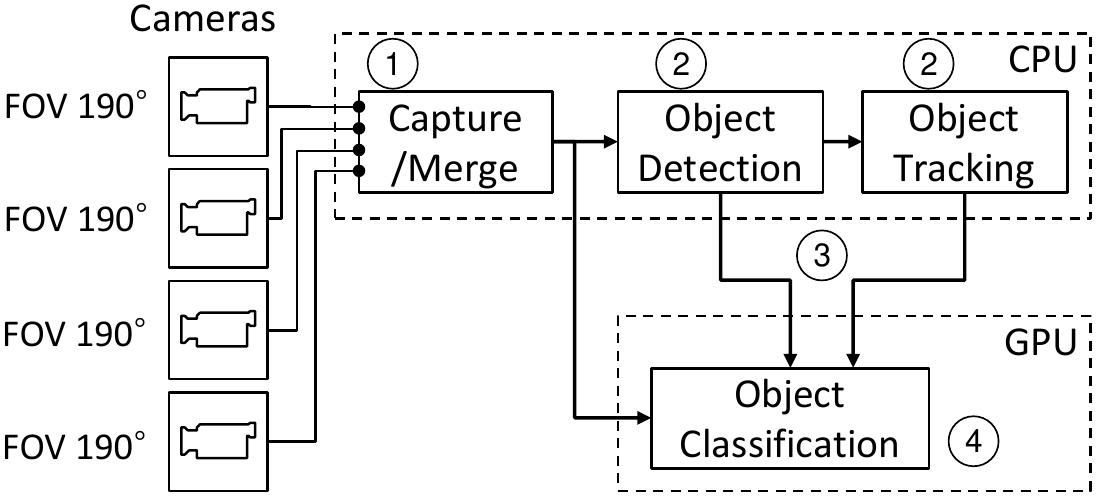}
	\caption{System Architecture: The modules are independent from each other and can be enabled individually.}
	\label{fig:sys_arch}	
\end{figure}

\begin{table}[b]
\begin{center}
\caption{System Specification of different platforms}
\label{table:platform_specs}
\begin{tabular}{|c|c|c|}
\hline
Platform & CPU & GPU/DSP\\
\hline
TDA2x & Dual-core 1.0-GHz & 2 C66x DSPs \\
        &  ARM A15 processor & \\
\hline
TX2 & Quad-core 2.0-GHz ARM A57 & Pascal GPU \\
        & Dual-core 2.0-GHz Denver 2 & (2 SMs 256 Cores) \\
\hline
x86 & 3.3-Ghz Intel I9 7900x & Titan Xp GPU \\
& & (60 SMs 3840 Cores) \\
\hline
\end{tabular}
\end{center}
\end{table}

In this evaluation, we examined the practical feasibility and efficacy of our algorithm using multiple platforms from different vendors. Figure~\ref{fig:surroundview} shows our surround view system prototype built using a TI TDA2x evaluation kit from Spectrum Digital~\cite{TI_TDA2x}\cite{TI_TDA2x_Kit} to capture data and evaluate our detection/tracking algorithm in real-life scenarios. We also integrated our algorithm on an NVIDIA TX2 embedded platform~\cite{NVIDIA_TX2} and an x86 desktop to compare the relative performance of our detection, tracking and classification modules across different platforms. Specifications of the different hardware we used are listed in Table~\ref{table:platform_specs}. We ran 64-bit Ubuntu~16.04 with OpenCV 2.4.17 for the detection/tracking module. CUDA 8.0+CuDNN 6.0, Tensorflow, Python 3, and OpenCV 3.0 were used for the classification module implementation. To have the best classification result, we applied MobileNet 1.0 model to our system. We discuss speed performance with different MobileNet model versions in Section~\ref{experiment_result}.

Figure~\ref{fig:sys_arch} captures our implemented system architecture. The different components of the system architecture are as follows:
\begin{enumerate}
    \item Capture/Merge: This component captures the images from the 4 fisheye cameras and merges them into a single quad view image as shown in Figure~\ref{fig:ROI}.
    \item Detection/Tracking: These modules take the quad view image as input and process them to detect the moving objects. The outputs from the modules are the indexes of the ROIs containing moving objects. 
    \item Detection Output Transfer: The output from detection/tracking modules are transferred to the classification module using socket communication.
    \item Object Classification: This module takes the quad view image and selected ROI indexes as input and processes them to classify the moving objects into different categories.
\end{enumerate}

\begin{table}[]
\begin{center}
\caption{Speed Performance Breakdown of Detection module between GPU and CPU on NVIDIA TX2}
\label{table:table_speed_comp_gpu_cpu}
\begin{tabular}{|c||c|c|c|}
\hline
Core & Job description & Run Time (ms) & Total (ms) \\
\hline
CPU & Diff of two frames & $2.98\pm0.21$ & $43.11\pm4.58$ \\
    & Wise multiple & $0.01\pm0.08$ & \\
    & ORB Feature Extraction & $23.97\pm1.68$ & \\
    & Sparse Optical flow cal. &  $15.15\pm2.53$ & \\
\hline
GPU & Diff of two frames & $5.08\pm0.0s$ & $38.13\pm2.39$ \\
    & Wise multiple & $1.01\pm0.08$ & \\
    & ORB Feature Extraction & $19.64\pm1.13$ & \\
    & Sparse Optical flow cal. & $12.10\pm1.13$ & \\
\hline
\end{tabular}
\end{center}
\end{table}

\subsection{Experiment Results}
\label{experiment_result}

Each algorithm module presented in Figure~\ref{fig:detection_tracking_module} and~\ref{fig:classification_module} was developed to be independent from each other. We measured the performance and accuracy for 4 different module combinations \begin{itemize}
    \item Experiment 1: Detection only
    \item Experiment 2: Detection and tracking
    \item Experiment 3: Classification only
    \item Experiment 4: Detection, tracking and classification
\end{itemize}
The metrics for Experiment 1 and 2 were compared to evaluate the improvement in accuracy on adding the tracking module. Likewise, the metrics from Experiment 3 and 4 were compared to evaluate the speed and accuracy trade offs when combining all the modules.

We collected video sequences at 1280x720 resolution at 30 fps using our surround view system equipped with 4 190\degree FOV fisheye cameras as shown in Figure~\ref{fig:surroundview}. We used 7 video sequences to create our training data. 1009 frames were selected from the video data and 12 pre-fixed ROIs were sliced into individual images. The images were labeled with presence information and the category of objects for training our MobileNet-based classification module. To create test data, we labeled 5 other video sequences with manually-annotated ground truths of object presence and target class in 12 ROIs. These training and testing data were obtained from various indoor and outdoor parking lots from local shopping centers and buildings in Pittsburgh, USA.

\begin{table}[b]
\begin{center}
\caption{System Performance on different platforms}
\label{table:table_speed}
\begin{tabular}{|c|c|c|c|c|}
\hline
Platform    & Detection & \makecell{Detection\\Tracking} & Classification & \makecell{Detection\\Tracking\\Classification} \\
\hline
x86 & 28fps & 25fps & 15fps & 23fps \\
\hline
TX2 & 21fps & 15fps & 9fps & 11fps \\
\hline
TDA2x & 20fps & 10fps & NA & NA \\
\hline
\end{tabular}
\end{center}
\end{table}

Table~\ref{table:table_speed_comp_gpu_cpu} shows a comparison of the execution times of the detection module between GPU and CPU on NVIDIA TX2. We observe that running the detection module on the GPU instead of the CPU does not result in a significant performance improvement. This is because the algorithms deal with sparse features instead of processing all the pixels in the image. Therefore, the detection and tracking module can be selected to run either on a GPU or a CPU as per system availability. We chose to pin the detection and tracking operation to the CPU. This allowed us to dedicate the GPU resource to the DNN-based classification algorithm to maximize its performance.

\begin{table}[b]
\begin{center}
\caption{Accuracy Performance on NVIDIA TX2}
\label{table:table_accuracy}
\begin{tabular}{|c|c|c|c|c|}
\hline
    & Detection & \makecell{Detection\\Tracking} & Classification & \makecell{Detection\\Tracking\\Classification} \\
\hline
Precision & 87.62\% & 72.33\% & 91.59\% & 92.59\% \\
\hline
Recall & 82.33\% & 91.22\% & 91.48\% & 89.36\% \\
\hline
\end{tabular}
\end{center}
\end{table}

Table~\ref{table:table_speed} shows the measured average frame per second (fps) of each algorithm module running on different platforms. Unfortunately, we could not evaluate the classification module on the TI TDA2x platform since it does not support CUDA~\cite{CUDA}. We notice that the overall performance of the integrated ‘detection, tracking, and classification’ approach is significantly better than the 'classification only' scheme. This is because fewer ROIs need to be processed by the classification module using the filtered ROI index information received from the detection and tracking module. We also observe that the ‘detection, tracking, and classification’ experiment was practically feasible on an embedded platform like the NVIDIA TX2 by dedicating the entire GPU resource to the classification module. The NVIDIA TX2 supports a feature inherent in integrated GPUs called zero-copy memory. Zero-copy memory eliminates the need for copying data to and from DRAM associated with the GPU. This allows both CPU and GPU components of GPU intensive programs to share memory space. This feature enables us to reduce the latency while transferring the captured input image from the CPU to the classification module on GPU.

To evaluate the accuracy of our algorithms, we defined the following metrics for detection/tracking and classification modules:
\newline
\newline
$recall= \frac{Correctly\ detected\ ROIs(True Positive)}{Total\ \#\ of\ ROIs\ with\ moving\ object(Ground\ Truth)}$
\newline
\newline
$precision=\frac{Correctly\ detected\ ROIs(True\ Positive)}{All\ Detected\ ROIs(True\ Positive+False\ Positive)}$
\newline
\newline

Our measured values for the precision and recall are shown in Table \ref{table:table_accuracy}. As we can see from the table, the addition of the tracking module improves the recall metric at the cost of less precision. Consider the case when a moving object comes to a stop at one of the ROIs. This object will not be detected in the absence of the tracking module. This ROI will be counted as a false negative in all the frames until the object starts moving again. The result is a low recall value. The addition of the tracking module does reduce the number of false negatives leading to a higher recall. But, we also see a reduction in precision. This happens in the scenario when the tracking module erroneously latches onto a non-moving object. This ROI then counts as a false positive for the subsequent frames resulting in a lower precision value. We believe that the overall compromise is however preferable for an autonomous vehicle system where we would rather have more false positives than false negatives. Another important takeaway is that only performing classification gives the best precision and recall values. However, without the detection and tracking modules, we will have to run the classification module for all ROIs of all frames which incurs a huge computational cost. In the future, with advances in hardware, it might be feasible to skip the detection and tracking modules to achieve better accuracy.

\begin{figure}[]
	\centering
	\includegraphics[width=0.8\columnwidth]{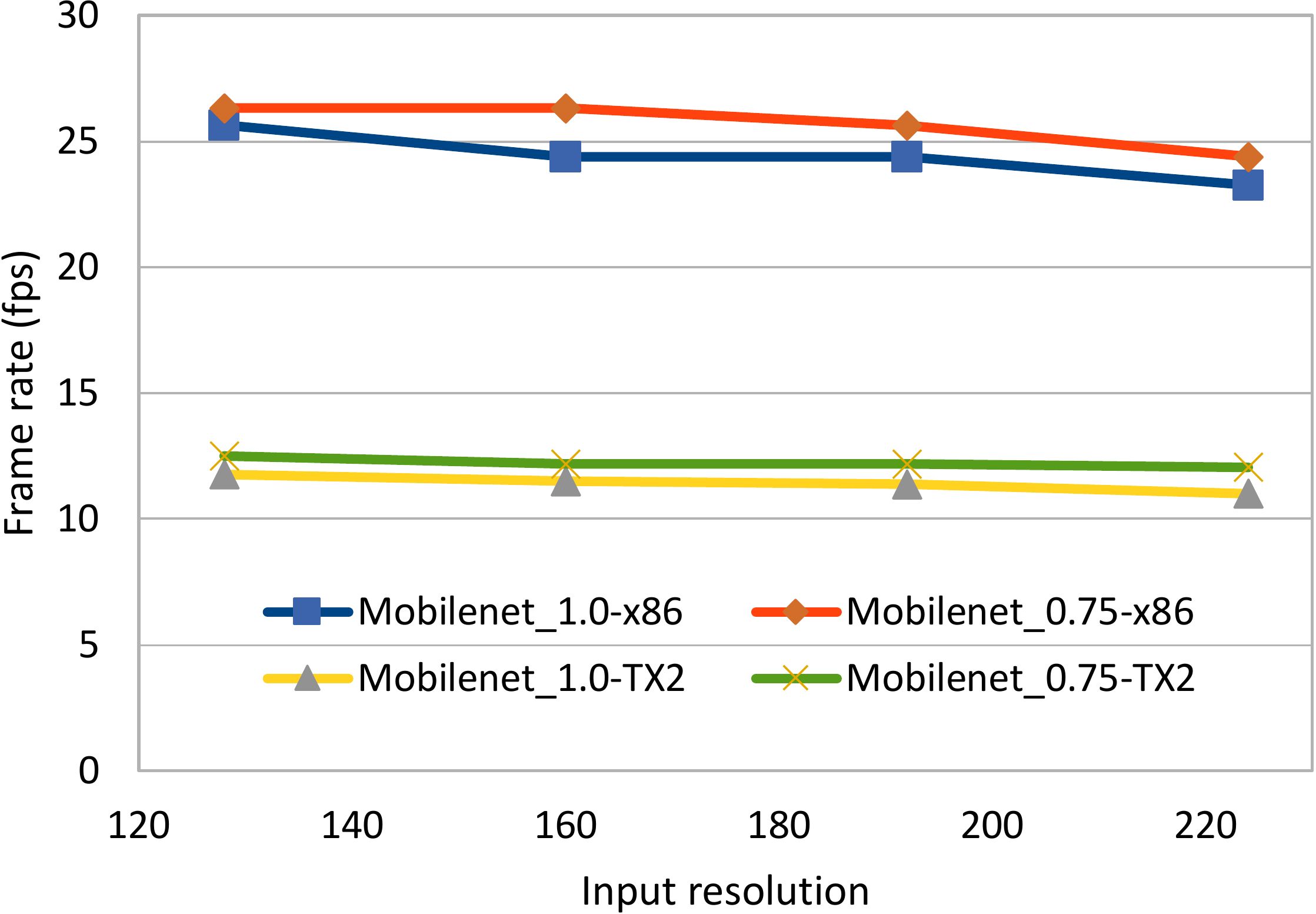}
	\caption{Speed vs. Resolution w.r.t. Different MobileNet version: MobileNet\_1.0 with 224x224 was applied to our system to achieve the best classification result. }
	\label{fig:mobilenet_speed}	
\end{figure}

Figure~\ref{fig:mobilenet_speed} shows the speed of different MobileNet models on x86 and NVIDIA TX2 with respect to the different input image resolutions. The appropriate MobileNet model can be chosen to fit the latency and size budget for the platform in use. We decided to use MobileNet\_v1\_1.0\_224~\cite{mobileNetBlog} to achieve the best accuracy in real-time.

The test result of our classifier was compared with Faster R-CNN~\cite{ren2015faster}. We used MobileNet as the backbone of the R-CNN network for fair comparison. The Faster R-CNN module was integrated only on the x86 desktop because it requires more computational resources than those provided by NVIDIA TX2. We trained a model using KITTI dataset~\cite{geiger2013vision} and measured the performance using our dataset. The system performance showed 15fps and the accuracy performance showed 60.5\% and 81.1\% for recall and precision, respectively. We observe a significant reduction in precision because the Faster R-CNN module fails to detect objects along the edges of the image where the distortion is severe. We expect that the accuracy will be improved using our own training data.

\section{Conclusion and Future Work}

We have presented a solution that can detect, track and recognize moving objects in pre-determined regions of interest in real-time with good accuracy. Our solution was evaluated on x86, TI TDA2X and NVIDIA TX2 platforms and their relative performance was measured for various combinations of active modules. The final solution was also deployed on a real car and extensively tested in real-world situations. The system performed with good accuracy and precision in these real-world tests. As the system was built from the ground up using a modular approach, the advantage of modularity was evident in the ease with which the breakdown of performance and accuracy measurements could be taken. In fact, these modules can be effectively added to any existing autonomous vehicle solution with minimum porting time and system overhead.

In the future, we plan to train the classification module to detect several additional categories. We would also like to add an ego-motion compensator to be able to use the detection and tracking module even when the vehicle is traveling at high speeds. Another next step is to integrate the entire solution with other existing algorithms on the NVIDIA Drive PX2 and deploy on our autonomous driving research vehicle.

\bibliographystyle{IEEEtran} 
\raggedright
\bibliography{IEEEabrv,mybibfile}

\begin{thebibliography}{10}
\providecommand{\url}[1]{#1}
\csname url@rmstyle\endcsname
\providecommand{\newblock}{\relax}
\providecommand{\bibinfo}[2]{#2}
\providecommand\BIBentrySTDinterwordspacing{\spaceskip=0pt\relax}
\providecommand\BIBentryALTinterwordstretchfactor{4}
\providecommand\BIBentryALTinterwordspacing{\spaceskip=\fontdimen2\font plus
\BIBentryALTinterwordstretchfactor\fontdimen3\font minus
  \fontdimen4\font\relax}
\providecommand\BIBforeignlanguage[2]{{%
\expandafter\ifx\csname l@#1\endcsname\relax
\typeout{** WARNING: IEEEtran.bst: No hyphenation pattern has been}%
\typeout{** loaded for the language `#1'. Using the pattern for}%
\typeout{** the default language instead.}%
\else
\language=\csname l@#1\endcsname
\fi
#2}}

\bibitem{SRX}
J.~Wei, J.~M. Snider, J.~Kim, J.~M. Dolan, R.~Rajkumar, and B.~Litkouhi,
  ``Towards a viable autonomous driving research platform,'' in
  \emph{Intelligent Vehicles Symposium (IV), 2013 IEEE}.\hskip 1em plus 0.5em
  minus 0.4em\relax IEEE, 2013, pp. 763--770.

\bibitem{wang2017pedestrian}
H.~Wang, B.~Wang, B.~Liu, X.~Meng, and G.~Yang, ``Pedestrian recognition and
  tracking using 3d lidar for autonomous vehicle,'' \emph{Robotics and
  Autonomous Systems}, vol.~88, pp. 71--78, 2017.

\bibitem{stereo}
N.~Bernini, M.~Bertozzi, L.~Castangia, M.~Patander, and M.~Sabbatelli,
  ``Real-time obstacle detection using stereo vision for autonomous ground
  vehicles: A survey,'' in \emph{Intelligent Transportation Systems (ITSC),
  2014 IEEE 17th International Conference on}.\hskip 1em plus 0.5em minus
  0.4em\relax IEEE, 2014, pp. 873--878.

\bibitem{bertozzi2015360}
M.~Bertozzi, L.~Castangia, S.~Cattani, A.~Prioletti, and P.~Versari, ``360
  detection and tracking algorithm of both pedestrian and vehicle using fisheye
  images,'' in \emph{Intelligent Vehicles Symposium (IV), 2015 IEEE}.\hskip 1em
  plus 0.5em minus 0.4em\relax IEEE, 2015, pp. 132--137.

\bibitem{cho2014multi}
H.~Cho, Y.-W. Seo, B.~V. Kumar, and R.~R. Rajkumar, ``A multi-sensor fusion
  system for moving object detection and tracking in urban driving
  environments,'' in \emph{Robotics and Automation (ICRA), 2014 IEEE
  International Conference on}.\hskip 1em plus 0.5em minus 0.4em\relax IEEE,
  2014, pp. 1836--1843.

\bibitem{choi2012realtime}
J.~Choi, ``Realtime on-road vehicle detection with optical flows and haar-like
  feature detectors,'' Tech. Rep., 2012.

\bibitem{sift}
D.~G. Lowe, ``Object recognition from local scale-invariant features,'' in
  \emph{Computer vision, 1999. The proceedings of the seventh IEEE
  international conference on}, vol.~2.\hskip 1em plus 0.5em minus 0.4em\relax
  Ieee, 1999, pp. 1150--1157.

\bibitem{surf}
H.~Bay, T.~Tuytelaars, and L.~Van~Gool, ``Surf: Speeded up robust features,''
  \emph{Computer vision--ECCV 2006}, pp. 404--417, 2006.

\bibitem{brief}
M.~Calonder, V.~Lepetit, C.~Strecha, and P.~Fua, ``Brief: Binary robust
  independent elementary features,'' \emph{Computer Vision--ECCV 2010}, pp.
  778--792, 2010.

\bibitem{orb}
E.~Rublee, V.~Rabaud, K.~Konolige, and G.~Bradski, ``Orb: An efficient
  alternative to sift or surf,'' in \emph{Computer Vision (ICCV), 2011 IEEE
  international conference on}.\hskip 1em plus 0.5em minus 0.4em\relax IEEE,
  2011, pp. 2564--2571.

\bibitem{baker2004lucas}
S.~Baker and I.~Matthews, ``Lucas-kanade 20 years on: A unifying framework,''
  \emph{International journal of computer vision}, vol.~56, no.~3, pp.
  221--255, 2004.

\bibitem{kristan2015visual}
M.~Kristan, J.~Matas, A.~Leonardis, M.~Felsberg, L.~Cehovin, G.~Fern{\'a}ndez,
  T.~Vojir, G.~Hager, G.~Nebehay, and R.~Pflugfelder, ``The visual object
  tracking vot2015 challenge results,'' in \emph{Proceedings of the IEEE
  international conference on computer vision workshops}, 2015, pp. 1--23.

\bibitem{alexnet}
A.~Krizhevsky, I.~Sutskever, and G.~E. Hinton, ``Imagenet classification with
  deep convolutional neural networks,'' in \emph{Advances in neural information
  processing systems}, 2012, pp. 1097--1105.

\bibitem{howard2017mobilenets}
A.~G. Howard, M.~Zhu, B.~Chen, D.~Kalenichenko, W.~Wang, T.~Weyand,
  M.~Andreetto, and H.~Adam, ``Mobilenets: Efficient convolutional neural
  networks for mobile vision applications,'' \emph{arXiv preprint
  arXiv:1704.04861}, 2017.

\bibitem{TI_TDA2x}
``{TDA2x SoC family},'' \url{http://processors.wiki.ti.com/index.php/TDA2x}.

\bibitem{TI_TDA2x_Kit}
``{TDA2x Vision Evaluation Module Kit},''
  \url{http://www.spectrumdigital.com/tda2x-vision-evaluation-module-kit}.

\bibitem{NVIDIA_TX2}
``{NVIDIA Jetson TX1/TX2 Embedded Platforms},''
  \url{http://www.nvidia.com/object/embedded-systems-dev-kits-modules.html}.

\bibitem{CUDA}
J.~Nickolls, I.~Buck, M.~Garland, and K.~Skadron, ``Scalable parallel
  programming with {CUDA},'' \emph{ACM Queue}, vol.~6, no.~2, pp. 40--53, 2008.

\bibitem{mobileNetBlog}
``{MobileNets: Open-Source Models for Efficient On-Device Vision},''
  \url{https://research.googleblog.com/2017/06/mobilenets-open-source-models-for.html}.

\bibitem{ren2015faster}
S.~Ren, K.~He, R.~Girshick, and J.~Sun, ``Faster r-cnn: Towards real-time
  object detection with region proposal networks,'' in \emph{Advances in neural
  information processing systems}, 2015, pp. 91--99.

\bibitem{geiger2013vision}
A.~Geiger, P.~Lenz, C.~Stiller, and R.~Urtasun, ``Vision meets robotics: The
  kitti dataset,'' \emph{The International Journal of Robotics Research},
  vol.~32, no.~11, pp. 1231--1237, 2013.

\end{thebibliography}

\end{document}